\documentclass[]{elsarticle}

\usepackage{hyperref}

\usepackage{caption} %改变图表标题

\usepackage{booktabs} %调整表格线与上下内容的间隔

\usepackage{longtable}%调用跨页表格
\usepackage{amsmath,bm}

\usepackage{numcompress}\biboptions{authoryear}

\begin{document}

\begin{frontmatter}

\title{Positive emotions help rank negative reviews in e-commerce }
\tnotetext[mytitlenote]{This work was supported by NSFC (Grant No. 71871006).}

\author{Di Weng}
\address{School of Economics and Management, Beihang University, Beijing, China.}
%\ead{di.W@buaa.edu.cn}

\author{Jichang Zhao}
\address{School of Economics and Management, Beihang University, Beijing, China.}
\address{Beijing Advanced Innovation Center for Big Data and Brain Computing, Beijing, China.}
\ead{jichang@buaa.edu.cn}


\begin{abstract}
Negative reviews, the poor ratings in postpurchase evaluation, play an indispensable role in e-commerce, especially in shaping future sales and firm equities.\ However, extant studies seldom examine their potential value for sellers and producers in enhancing capabilities of providing better services and products.\ For those who exploited the helpfulness of reviews in the view of e-commerce keepers, the ranking approaches were developed for customers instead. To fill this gap, in terms of combining description texts and emotion polarities, the aim of the ranking method in this study is to provide the most helpful negative reviews under a certain product attribute for online sellers and producers.\ By applying a more reasonable evaluating procedure, experts with related backgrounds are hired to vote for the ranking approaches.\ Our ranking method turns out to be more reliable for ranking negative reviews for sellers and producers, demonstrating a better performance than the baselines like BM25 with a result of 8\% higher.\ In this paper, we also enrich the previous understandings of emotions in valuing reviews. Specifically, it is surprisingly found that positive emotions are more helpful rather than negative emotions in ranking negative reviews.\ The unexpected strengthening from positive emotions in ranking suggests that less polarized reviews on negative experience in fact offer more rational feedbacks and thus more helpfulness to the sellers and producers.\ The presented ranking method could provide e-commerce practitioners with an efficient and effective way to leverage negative reviews from online consumers.
\end{abstract}


\end{frontmatter}

%%\linenumbers

\section{Introduction}
\qquad As Internet technology develops rapidly in recent decades, e-commerce has grown tremendously with the technology which the business based on, especially in China, where online shopping prevails among from city to countryside.\ According to CNNIC (China Internet Network Information Center), there were more than 710 million people who shop online in the first quarter of 2020 and generate abundant online reviews.\ Those online reviews contain real-time feedbacks from consumers, mainly in formats of texts and ratings.\ Along with online feedback mechanism, online reviews can help build up trust and foster cooperation between strangers \citep{dellarocas2003digitization}.\ Customers are in fact getting dependent on online reviews by previous buyers to value whether it is proper to purchase the products that online retailers provide.\ Even more inspiring, online reviews are posted after experiencing products and accordingly these feedbacks from different users can further be aggregated to offer insights of upgrading services and refining production for both sellers and producers involved in e-commerce.

\qquad Online reviews play profound roles in the business model of e-commerce.\ They can affect online sellers and customers and even producers significantly.\ Based on personal experience, online customer reviews are user-oriented to be helpful for less-sophisticated customers to target the best-matched products \citep{chen2008online}.\ The description sentences about the products help potential customers to make decisions \citep{schindler2012perceived}. Even the rating itself is influenced by previously posted reviews~\citep{moe2011value}.\ Most online retailers also believe their performance is diminished due to their unable decipher or reliably assess how online customers use the informational cues from the reviews and conversations \citep{bonnet2011transform}.\ Online user-generated reviews are also of importance to business performance in extensive circumstances like tourism \citep{ye2011influence}. \ It can even be profitable for the sellers to delay the availability of consumer reviews \citep{chen2008online}.

\qquad In particular, the complaints in negative reviews may include useful and profitable information for both of online retailers and producers to make adjustments and improvements.\ Positive and negative reviews are correlated with the box office revenue, while it was underscored that the impact of negative reviews to hurt the performance of box office is relatively more significant than that of positive reviews to bolster the performance \citep{basuroy2003critical}.\ The similar result occurs in online book reviews, where the power of a negative review is more than that of a positive review when deciding whether the sales increase or decrease \citep{chevalier2006effect}.\ In accordance with these, being online word-of-mouth in e-commerce, negative reviews indeed impact the future sales of products~\citep{chevalier2006effect} and reshape the values of firm equity~\citep{luo2009quantifying}. Therefore, negative reviews and comments are thought to be more influential and helpful than positive reviews \citep{basuroy2003critical,forman2008examining,berger2010positive,chen2013temporal,yin2014anxious,huang2017social}.\ The underlying information in negative reviews therefore could be valuable in helping online sellers and producers enhance capabilities of providing services and products, which makes negative reviews worth being investigated and in particular ranked from their views.

\qquad In the meantime, positive and negative emotions are important parts in online reviews, either positive or negative. Except the descriptive texts towards factual details in product experience, rich emotions carried in reviews also signal feelings and attitudes of consumers in postconsumption evaluation.\ Positive emotions are more prevalent in positive reviews, e.g. 5-star ratings. On the contrary, negative emotions are more dominant in negative reviews, e.g. poor ratings with 1-star.\ Nevertheless, this polarity phenomenon does not necessarily indicate that there is no positive emotions in negative reviews or vice versus.\ It is widely investigated by previous researchers of the relationship between emotions and helpfulness of reviews.\ Negative reviews are more helpful when perceived by people, yet this helpfulness decreases when the emotion intensity is too extreme\citep{lee2017roles}. While reviews that contain extreme opinions were also found to be more helpful than those with mixed or neutral \citep{cao2011exploring}.~\cite{schindler2012perceived} found that positive information may lead readers to make further consideration of the product, but too much positiveness may cause suspicion of the reviewer's motives. However, recent efforts suggest that positive emotion expressions in negatively rated reviews are perceived less useful, also less frequent comparing with negative emotion expressions\citep{kim2020roles}. These disputed evidences imply the context-dependence of emotions in determining review helpfulness and suggest the necessity of considering contexts in evaluating their impact in ranking reviews. Specifically, from the unique view of online sellers and producers, whether emotions impact helpfulness of negative reviews and if so, to what extant sill remain unclear and deserve further explorations.

\qquad \ Given a mass amount of online reviews, it is unrealistic for either side to go through all of them.\ Nevertheless, effective methods for ranking reviews are not compatible with the advancement of e-commerce, especially in the perspective of sellers and producers.\ Former studies focused on consumer reviews' impact on potential buyers.\ For example, the impact of reviews on sales and other consumer's perceived helpfulness can be predicted by predictive modeling techniques such as econometric regressions and machine learning algorithms~\citep{ghose2011estimating}.\ External information in product description data and customer question answer data can also help improve the prediction of the helpful reviews \citep{saumya2018ranking}.\ However, in evaluations of these ranked reviews, those who make votes were predominantly supposed to be future buyers.\ That is to say, these ranking solutions are more based on consumers rather than sellers and producers. Besides, existing studies seldom illustrate the helpfulness of negative reviews which is absolutely very important from the perspective of sellers and producers. In addition, the debate over which kind of emotions in negative reviews is more useful further calls for careful examination in context-dependent evaluations. Due to these concerns, sellers and producers oriented ranking algorithms over negative reviews, aiming at targeting influential feedbacks on defeats of services and products, should be established in fill the vital gap in existing studies.

\qquad No question that negative reviews are the key to making e-commerce better, in particular for online sellers and producers.\ We thus develop an efficient ranking method to help e-commerce sellers and producers to improve their business by effectively retrieve information about what exactly customers are complaining about their products relating to attributes such as logistics, quality, marketing, and customer service.\ Using 65, 702, 406 negative reviews of 1-star ratings from JD.com, one of the largest e-commerce platforms in China, we first train word vector models to infer the embeddings and measure semantic similarities.\ Then we calculate the emotion polarities of every single review, combining them with the semantic distances of different attributes in order to target the most useful negative review in an attribute resolution.\ More importantly, experts with related backgrounds are instructed to value the helpfulness of negative reviews from the very view of online sellers and producers. Through these unique but practical settings, it is found that positive emotions are relatively more helpful than negative emotions in ranking negative reviews, which means that negative emotions are slightly less critical than that of the positive.\ The helpfulness rates of our ranking method are much higher than the baseline methods.\ Interestingly, it is also revealed that when giving higher weights for negative emotions, the ranking results turn out to be more fluctuating than not, meaning that highly negatively polarized emotions could undermine the helpfulness of negative reviews.\ Our results not only supplement strong evidence to enrich the existing understandings on contributions of emotions in valuing reviews, but also provide the e-commerce practitioners with an effective and efficient way to get direct and valuable feedbacks from their customers.

\section{Literature Review}
\qquad There are two streams of previous literatures that primarily relate to our study.\ The first one is the importance of online reviews, negative reviews to be precise.\ It has been broadly stated that negative reviews are relatively more helpful than that of the positive.\ For example, people would be more interesting in the blemishes of the products they are going to purchase.\ And such information often comes with negative emotions. One of the main directions of extant studies in this stream mainly relates to what influence the reviews can bring for online buyers and online sellers or the role the reviews act in the whole market.\ Another part of the previous work concerns the ways to evaluate reviews; this part includes information retrieval methods and ranking methods.\ In this section, we review the researches that make significant contributions and point out the imperfectness of existing efforts that motivates the present study.

\subsection{Importance of Reviews}
\qquad Online reviews are of great importance in the business model of e-commerce.\ \cite{dellarocas2003digitization} examined the difference between internet-based feedback mechanism and traditional word-of-mouth networks.\ They pointed out that it is the online review and the internet-based feedback mechanism that makes strangers to cooperate possible.\ Online reviews also serve as the way to build confidence and facilitate cooperation in online marketing.\ Most online shopkeepers attributed the poor performance to the unable effective analysis of, or the reliable assessment of online customers use of online reviews \citep{bonnet2011transform}.\ As for the customer reviews in online shopping platforms, especially negative reviews, they tended to contain much information regarding customers experience and products blemishes \citep{chen2013temporal,mudambi2010makes}.

\qquad Thus, tremendous efforts have been dedicated on the impact of customer reviews, either positive or negative.\ \cite{chevalier2006effect} examined the effect of consumer reviews about books on Amazon.com and Barnesandnoble.com.\ Based on their model, both cross-sectional and differences-in-differences analysis found that positive reviews would help improve book sales online.\ Apart from that, they also concluded that the decreasing effect of sales resulting from an incremental negative review is more powerful than the increasing impact resulting from an additional positive review.\ \cite{chen2008online} even developed a normative model to help sellers change marketing strategy in response to consumer reviews.\ As they expatiated, customer review could assist customers in recognizing the best-matching product.\ For example, online reviews would help less-sophisticated consumers in matching most-suited products.\ \cite{ludwig2013more} found the tapering-off effect of positive affective content, but not occurred in negative affective content, which demonstrated that the influence of positive content and negative content are asymmetric.\ \cite{berger2010positive} investigated the effects of negative reviews and found that negative publicity has positive influences on products, such as increasing the awareness of an unknown product.\ As a result, they could unexpectedly increase the chance of being purchased. In terms of dividing the products into two parts, search goods, and experience goods,\ \cite{mudambi2010makes} demonstrated that the extreme reviews are less helpful than moderate reviews for the former products; on the contrary, extreme reviews are more helpful for the latter.\ \cite{willemsen2011highly} found negative information in reviews are more paramount in making purchase decision, while only for experience products rather than search products, comparing to positive information; conversely, the latter products are more dominated by positive information.\ \cite{huang2017social} found that the social network integration improves the reviews' quantity but decreases the quality, while emotional language increases and cognitive language declines.

\qquad Emotion polarities existing in online reviews significantly affect the helpfulness of the reviews. \cite{forman2008examining} examined the relationship between reviews and sales, concluded that moderate book reviews are less helpful than extreme reviews.\ \cite{yin2014anxious} also held the view that emotional and positive reviews are relatively less helpful perceived by customers compared with negative reviews.\ However, \cite{duan2008online} stated that online users' reviews do not influence the movies box office significantly, nor do the online reviews affect the purchase decision made by customers.\ A further study of the reason that positive reviews are less helpful was done by \cite{chen2013temporal}.\ Positive reviews tend to be more attributed to the reviewer rather than product experience, which is more related to negative reviews.\ Meanwhile, their research found that temporal cues could decrease negativity bias, to increase the value of positive reviews. \cite{yin2014anxious} further showed that emotional expression in descriptive sentences could harm the perceived value of online reviews. In the contrary, \cite{cao2011exploring} comprehensively investigated factors that influence helpfulness votes of online reviews using text mining and revealed that semantic characteristics are more helpful in affecting the received helpfulness votes comparing with stylistic and other characteristics.\ They also found that extreme reviews can get more helpful votes than compound or neutral reviews. Though contrary to each other, these results of emotions on review helpfulness imply the very essence of context-dependence in how emotions determine the usefulness of online reviews.  

\qquad Not only with reviews in online shopping platforms, the domains of online word-of-mouth are also expanded to include online social media such as Twitter.\ It is demonstrated that positive word-of-mouth is more cognitive, while that of negativeness is more emotionally based \citep{sweeney2014factors,verhagen2013negative}.\ For its emotionally based characteristics, negative word-of-mouth would be transmitted very fast among people \citep{de2008word,sweeney2012word}.\ With the help of internet and social platforms, online word-of-mouth could engage potential customers and customers, the latter could launch a revenge campaign if the company disappoint them which may eventually developed into a brand disaster if not handle well \citep{bach2012online,perkins2009power}.\ Actually, negative reviews in reviews are more important from the perspectives of sellers and producers.\ The re-examination of the helpfulness of negative reviews, in particular to revalue their contributions in reflection and improvement of sellers and producer, is accordingly necessary. Besides, lacking of negative reviews, especially the impact of positive emotions on the helpfulness of negative reviews in existing studies suggests further explorations. The incongruence in extant results also implies the needs to re-examine the impact of emotions on negative reviews in the context of voting helpfulness by sellers and producers.

\subsection{Existing Ranking Methods}
\qquad Given the massive consumers in online shopping, feedbacks from them can be tremendous in amount and how to display historical reviews for future consumers thus become a tricky problem. More importantly, how to efficiently and effectively extract insightful information that beneficial to pinpoint and resolve defects in services and products do matter in the very essence to sellers and producers.\ Here ranking reviews refers to giving the numerical order of helpfulness of a set of reviews in a descending way.

\qquad The most classical and influential ranking method in information retrieval is BM25, developed by \cite{robertson1994some}.\ Given its importance and widely using \citep{robertson1995okapi}, we employ BM25 as the preliminary baseline in this study.\ \cite{ghose2011estimating} firstly combined econometric, text mining, and predictive modeling techniques to estimate the helpfulness and economic impact of online reviews.\ They predicted perceived helpfulness by Random Forest based classifiers.\ However, the focus of their study was the text features that mattered sales and perceived usefulness, not including emotional polarities which are of great importance in conveying customers' attitudes.\ \cite{saumya2018ranking} predicted the helpfulness of the online reviews using random-forest classifier and gradient boosting regressor combined with cosine similarity.\ Their findings indicated that features from product description data and customer question-answer data could improve the prediction accuracy.\ Using Support Vector Regression, \cite{hsu2009ranking} ranked comments to promote high-quality social comments and filter out low-quality comments.\ However, these ranking methods are designed and implemented from the view of consumers and the evaluation by top-$n$ is easy to confuse the consumers given similar reviews.\ Here, the top-$n$ rate is the percentage of the top $n$ golden helpful reviews, which could be valued and screened by experts, that are also ranked among the top $n$ by a ranking method. Higher rate accordingly stands for more consistency of the method with human judgments.

\qquad In order to find the most helpful review of a certain product for potential customers' decision, \cite{liu2008modeling} proposed a nonlinear model based on radial basis function to predict helpfulness of online reviews.\ \cite{martin2014prediction} demonstrated that the emotionality would help in predicting online reviews helpfulness by both emotion lexicon and supervised classification.\ A model architecture was developed for word vector representations with a low computational cost, which can be used to the word by a particular vector, called word2vec \citep{mikolov2013efficient,le2014distributed}.\ \cite{kusnerWordEmbeddingsDocument2015} further extended the word vector of every single non-stop word from two documents to measure the distance. \cite{pennington2014glove} introduced the word-word cooccurrence matrix from the corpus, the GloVe(Global vectors for word representation) can better learn the context. In terms of embedding review texts into vectors, these established models can offer promising ways of combing descriptive sentences with emotion polarities in ranking reviews.

\qquad In conclusion, most of the existing researches held the view that emotional factors in online review can affect the perceived helpfulness.\ Also, positive reviews are relatively less beneficial comparing with negative reviews.\ However, such results were developed based on the view of customers rather than that of online shop keepers such as sellers and producers.\ Moreover, there is still one problem remaining in the evaluation method in those kinds of experiments.\ When asked to find the most helpful reviews from several choices, reviewers would be confused to discriminate the most helpful reviews from other slightly less helpful ones.\ The reviewers are also less convincing because lack of related backgrounds, in particular in retail or production.

\qquad So in this paper, we are going to develop a ranking method for online sellers and producers to easily fetch the most helpful reviews relating to a specific defect of the services and products.\ Combining with emotions in the sentence and word embedding, our ranking method will provide online sellers with an effective tool to figure out what exactly their customers are complaining about so as to improve their service and products.\ More importantly, we ask experts with related backgrounds to mark all helpful reviews given specific attributes then calculate helpful rates of different methods to avoid the hard-to-discriminate dilemma.

\section{Research Methodology}
\subsection{Data Collection and Data Clean}
\qquad We collected online reviews from JD.com, one of the most popular Amazon-like online shopping platforms in China, which owns the largest fulfillment infrastructure.\ It is reported that JD.com have more than 360 millions of annual active customers with a 82.9-billion dollars net revenue in 2019\footnote{Data from https://corporate.jd.com}.\ As a comparison, Amazon earns a 280-billion dollars net sales\footnote{Data from https://s2.q4cdn.com/299287126/files/doc\_financials/2020/ar/2019-Annual-Report.pdf} with 150-million Prime members\footnote{Data from https://www.statista.com/statistics/829113/number-of-paying-amazon-prime-members} in 2019.\ We used a web crawling technique to collect reviews from JD.com.\ The returned JSON data contained a field named ``score" denoting whether the current review was a negative comment, i.e, the ones with score 1 were negative reviews. The collected data were categorized into different kinds of products and stored by products.

\qquad A total of 117, 046, 285 negative reviews were collected from JD.com.\ The data collection was done for 246 categories including laptop, phone, refrigerator and other products with 2, 523, 212 different specific products.\ In this paper, we select laptop and phone as investigation targets. Because JD.com only sold computer, communication and consumer electronics when it was established.\ Those products are therefore the most popular items in JD.com, which have many purchasers and abundant reviews for investigation. Meanwhile, narrowing down the product types could also help low the labor cost of evaluation. We comprehensively collected detailed data fields for each online review, which greatly facilitate the following presentation of ranking approaches.\ Note that the selection of product types would not essentially effect the implementation of our proposed ranking methods, since the data fields considered in the ranking is independent to product types. And it is anticipated that the presented ranking method can be easily extended to other product categories.

\qquad The collected online negative reviews contained some irrelevant characters such as emoticons and URLs.\ We removed all these junk information from every negative review with python using regular expressions \citep{baeza1999modern}.\ With the help of Jieba (a Chinese word segmentation python module), we cut all of the 824GB negative reviews from JD.com into terms using our own stopping word dictionary, which contains 1589 special phrases. In the meantime, we filtered out the too short negative reviews whose lengths were less than 5 terms after the segmentation, which is a good threshold after several extracting tests.\ There are ultimately 65, 702, 406 negative reviews after the data clean.

\subsection{Problem Definition and Method Basis}
\qquad The problem setting in this paper is that finding the most helpful negative reviews from the perspective of sellers and producers given a set of negative reviews from an e-commerce platform (e.g., JD.com).\ Those reviews could help sellers and producers make adjustments and improvements of existing problems in services and products.\ To measure the distance between different words or product attributes, the word embedding is a good way to represent the word in the form of vector which can map into a specific and universal space.\ Common implementations of word embedding including one-hot, word2vec\citep{mikolov2013efficient, mikolov2013distributed}, GloVe\citep{pennington2014glove} and BERT\citep{devlin2018bert}.\ The one-hot word embedding uses sparse matrix to represent every word or sentence which is the basic way to transfer the word into a vector space.\ Although it has contained the co-occurrence information, the sparse vector is not compatible when there are plenty of words in the corpus as the dimension increases as well.\ The one-hot word embedding can not store the information about word similarity either.\ Word2vec uses a neural networks for word embedding, which defines the context with a window.\ However, GloVe further advances the word2vec using the whole corpus to create the word-word co-occurrence matrix, which means that it can better learn the features such as word analogy, word similarity from the corpus.\ BERT is a pre-trained model using bidirectional transformer with some tricks such as masked language model(MLM) and next sentence prediction (NSP).\ However, the performance of BERT on short texts, such as the negative reviews, are seldom discussed and investigated\citep{yu2019improving}.\ Besides, there are typically hundreds of millions parameters to optimize when training a BERT model, which makes a higher demand of computing resources\citep{devlin2018bert} and might undermine the efficiency of the ranking.\ When deciding the word embedding method to measure the similarity between given attribute words (topics such as quality, logistic, marketing, consumer service and so on) and negative reviews, we compare the top-$n$ rate of GloVe and word2vec using a small size of topics before applying to the larger size of topics in this paper.

\qquad Accordingly, we trained a GloVe model and a word2vec model using the corpus of negative reviews after the data clean, segmentation and filter.\ Those corpora contained word vectors that can measure semantic similarity between different words which occur in the corpus.\ Usually, the similarity is known as cosine distance or cosine similarity, which is calculated as
\begin{equation}
Similarity({v}_{w1},{v}_{w2})=\frac{{v}_{w1}\cdot{v}_{w2}}{\|{v}_{w1}\| \cdot\|{v}_{w2}\|},
\label{cosine_similarity}
\end{equation}
where ${v}_{w1}$ is the first word's vector from the GloVe or word2vec embedding and ${v}_{w2}$ is the second word's vector from the GloVe or word2vec embedding.\ Particularly, ${v}_{w1}$ and ${v}_{w2}$ have the same dimension.

\qquad The emotion lexicon is an important part of sentiment analysis.\ With such dictionaries, the emotional polarity can be directly captured from the given texts through occurrences of corresponding emotional terms.\ These dictionaries including ANEW words \citep{bradley1999affective}, SentiWordNet \citep{baccianella2010sentiwordnet} and the LIWC dictionary \citep{pennebaker2003psychological}.\ Here, we used collected reviews (not only the negative ones but also the positive ones with rating scores of 5) to create the emotion dictionary with a recursive method. \ Specifically, for negative dictionary, we started from the several man-select seed words, which contained both negative and positive terms.\ Then we counted every non-seed word's co-occurrences with the seed words.\ The negative and positive seed words were counted separately.\ The negative ratio is then calculated as
\begin{equation}
NegativeRatio=\frac{{n}_{n}}{{n}_{p}},
\label{negative_ratio}
\end{equation}
where ${n}_{n}$ denotes the non-seed word's co-occurrences of the negative seed words, ${n}_{p}$ denotes that of positive seed words.\ The we sorted the $NegativeRatio$ of every non-seed word in a descending order.\ Three well-instructed coders, who are experienced online buyers and reviewers in JD.com, were asked to jointly determine whether non-seed words could be added into the negative seed words.\ So as to add all of the non-seed words with high values of $NegativeRatio$ as the negative seed words.\ Those with high values of $NegativeRatio$ yet not be added into negative seed words were marked as not-seed words, which means we do not judge whether they could be negative seed words in next iteration.\ Recursively, all potential negative seed words were screened out of the non-seed words until all non-seed words' $NegativeRatio$ approaching 0.5, which denotes current words' has equal co-occurrence possibilities with both negative and positive seed words.\ Similarly, as for the positive seed words, the positive ratio is calculated as
\begin{equation}
PositiveRatio=\frac{{n}_{p}}{{n}_{n}},
\label{positive_ratio}
\end{equation}
and all potential positive seed words were marked out of the non-seed words until all non-seed words' $PositiveRatio$ approaching 0.5. Ultimately we built two Chinese emotion lexicons with the first one composed by 1, 589 unique negative words and the second one composed by 1, 135 unique positive words. Both lexicons can be publicly available at \url{https://doi.org/10.6084/m9.figshare.12327680.v1}.

\qquad Our goal is to find the most useful comment given specific topic word or phrase for online shop keepers.\ For example, an online seller would like to find out what exactly the customers complaining about concerned with the products' quality, so that he or she could make improvements to increase the competitiveness. More generally, those topic words and phrases should be semantically close to product attributes and customers' experience.\ Given one Chinese review, we firstly segmented it with Jieba and then used positive and negative emotion lexicons to measure the emotional polarity $e_{n}$ of the review as
\begin{equation}
e_{n}=\frac{p-n}{p+n},
\end{equation}
where $p$ is the number of positive words in the review text and $n$ is the number of negative words.\ The emotional polarity $e_{n}$ has a range of $[-1,1]$. And $e_{n}>0$ stands for positively polarized reviews while $e_{n}<0$ represents the negatively polarized reviews. Note that if $p+n=0$, i.e., there is no emotion words in the negative review, it will be treated as neutral with emotional polarity equals 0.

Given a topic word that represents a certain attribute of service or product, we used the GloVe or word2vec model to embed it into a word vector $v_{s}$.\ If given a topic phrase or review, we segmented it first then calculate the mean vector of all the words it contains.\ The similarity of the topic word or phrase and the review $C_{s}$ could be further measured as
\begin{equation}
C_{s}=\frac{{v_{s}}^{T} \cdot v_{p}}{\|v_{s}\| \cdot\|v_{p}\|},
\end{equation}
where $v_{s}$ is the vector represents the topic word or phrase, and $v_{p}$ is the mean vector of embeddings of the words that comprise the review.
	
\qquad Further, we introduced the $Sigmoid$ function which is very common in neural network models.\ By integrating the emotion polarity and the $Sigmoid$ function, the closer the absolute value of the emotion polarity is to 1, the less the increase in the emotion polarity caused by the increase of the same emotion word.\ The $Sigmoid$ function could soften the impact of the extreme emotions found in online reviews and smooth the fluctuation of emotion polarity.\ We also introduce three variations of $Sigmoid$ function named $inverseSigmoid(iSigmoid)$, $moutainSigmoid(mSigmoid)$ and $inverseMoutainSigmoid(imSigmoid)$ to perform a thorough investigation.\ The adjusted emotion polarity $e_{c}$ could be respectively calculated as
\begin{equation}
sigmoid:\qquad e_{c}=\frac{1}{1+e^{-e_{n}}},
\end{equation}
\begin{equation}
iSigmoid:\qquad e_{c}=\frac{1}{1+e^{e_{n}}},
\end{equation}
\begin{equation}
mSigmoid:\qquad e_{c}=\left\{\begin{array}{ll}{\frac{1}{1+e^{e n}}} & {e_{n}\geq0} \\ \\\ {\frac{1}{1+e^{-e_{n}}}} & {e_{n}<0}\end{array}\right.,
\end{equation}
\begin{equation}
imSigmoid:\qquad e_{c}=\left\{\begin{array}{ll}{\frac{1}{1+e^{-e_{n}}}} & {e_{n}\geq0} \\\\ {\frac{1}{1+e^{e_{n}}}} & {e_{n}<0}\end{array}\right..
\end{equation}
Then the rank score of every review given specific attribute could be calculated as
\begin{equation}
r_{s}=c_{s} * e_{c},
\label{eq:rankscore}
\end{equation}
where $c_{s}$ demonstrates the similarity of the selected review and the given attribute, which can be derived in the semantical level, while $e_{c}$ represents the emotion polarity of the review and it could be adjusted by diverse Sigmoid functions.\ In this ranking method, the review with highest rank score not only needed to be highly semantically related to the specified attribute, but also needed the high emotion polarity adjusted by given functions ($Sigmoid$ and three variations).

\qquad We first use tradition evaluating methods to test proposed ranking methods' outcome in different word embedding methods, word2vec and GloVe.\ There are 33 attribute phrases in both of laptop and phone.\ Three experts with backgrounds of management are instructed to rank the reviews extracted using different methods with the best result.\ Here we merge same reviews if different methods extract the same one.\ There are ten different methods including Glove, word2vec, both of the word embedding models with Sigmoid emotion function and its three variations as well as the baseline method BM25.\ Unlike word embedding based methods, the rank of BM25 totally depends on the BM25 index value without any emotion intervention.\ After the ranking process, we calculate every methods' top-$n$ rate.\ The average correct rate is calculated as well, which uses every method to sort all reviews that has been ranked by three experts to be compared with its man-selected order. Based on this result, we then determine the better word embedding model and apply it to a larger test set to perform a further evaluation. 

\begin{figure}[!ht] 
    \begin{minipage}[t]{1\linewidth}
    \centering 
    \includegraphics[scale = 0.4]{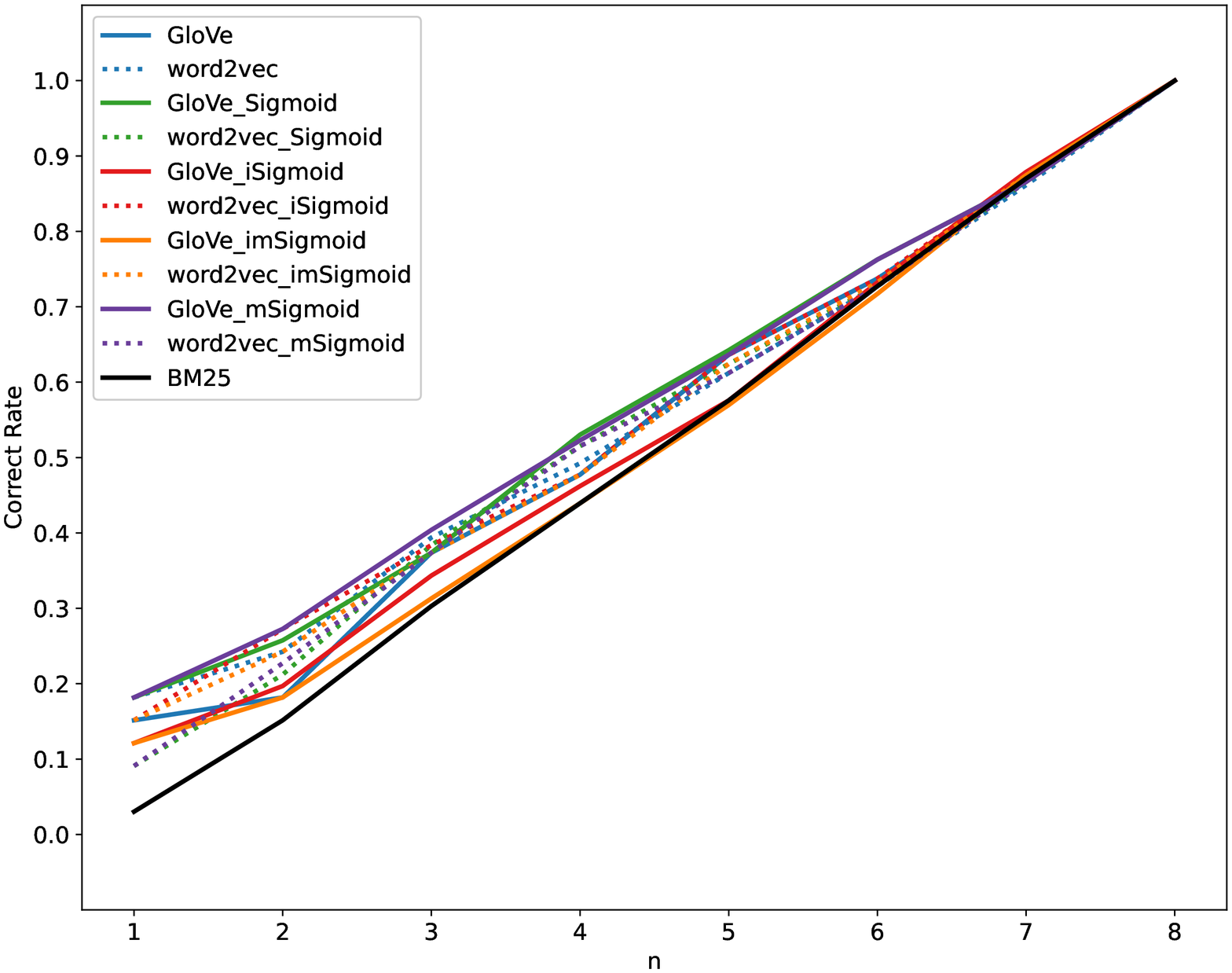}
    \caption*{(a)top-$n$}
    \end{minipage} 
    
    \begin{minipage}[t]{1\linewidth}
    \centering 
    \includegraphics[scale = 0.33]{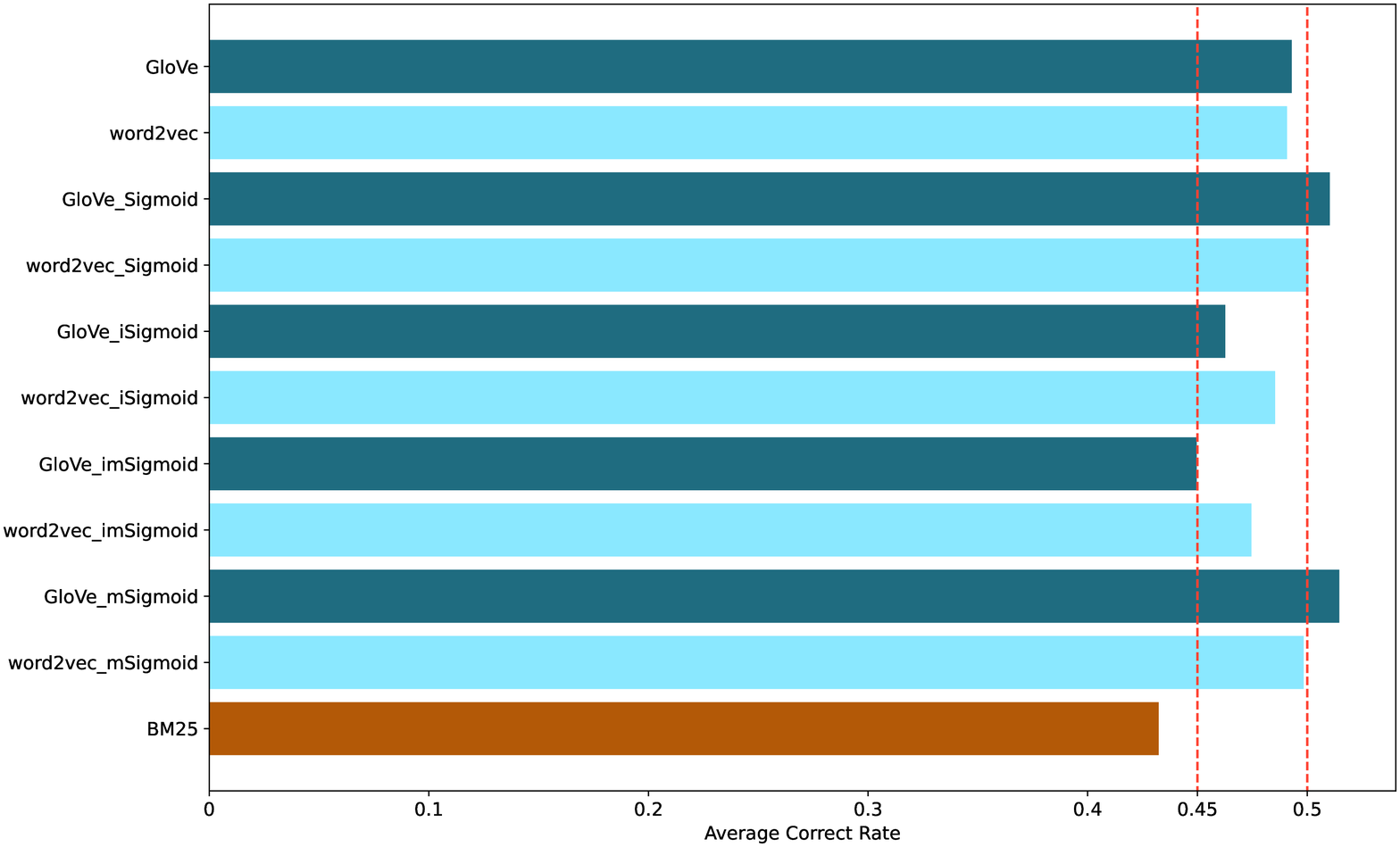} 
    \caption*{(b)Average Correct Rate} 
    \end{minipage} 
    \caption{Top-$n$ and Average Correct Rate and of GloVe-based and word2vec-based methods}
    \label{fig-pretest}
\end{figure} 

\section{Results and Discussions}
\subsection{Results}
\qquad The result of top-$n$ using 33 attribute phrases is shown in Figure \ref{fig-pretest}.\ We found that word2vec outperforms GloVe when $n$ is small other than GloVe\_$mSigmoid$ and GloVe\_$Sigmoid$.\ When it comes to the average correct rate, for the methods of word embedding and word embedding with $iSigmoid$ and $imSigmoid$, those methods have nearly 50\% of correct rate which outperform rest of the methods with lower average correct rate closing to 45\%.\ All methods using word embedding are better than BM25 which means that the word embedding is effective to achieve our goal, providing valuable reviews for online sellers or producers given specific attribute words or phrases.\ Among the methods of higher average correct rates, the results of methods using GloVe are better than that of using word2vec.\ While among the methods of lower average correct rates, the results of methods using GloVe are worse than that of using word2vec.\ According to the results of this 33 attribute phrases ranking, GloVe not only contains the best method in top-$n$ and also outperforms word2vec in average correct rate.\ Thus We choose GloVe as the word embedding model in the further evaluation.

\qquad Our primary purpose of the further evaluation is to screen the most proper method for online sellers and producers to target helpful feedbacks and make improvements of their online commerce.\ We thus selected 209 and 200 attributes in phone and laptop to extend the test set, separately and hire 6 experts with management backgrounds to evaluate the helpfulness of every review that extracted using different ranking methods.\ Specifically, given a attribute descriptive word, each ranking method was run to retrieve the one with the highest rank score.\ All experts were instructed to mark whether the review was helpful from the point of online sellers and producers given the specified service or product attributes.

\qquad After aggregating the evaluation of the experts, the helpfulness rate of each method was calculated, which is shown in Table \ref{helfulnessrate}.\ We selected BM25 and GloVe as the baselines.\ Overall, the simply implementing GloVe gets the worst outcome among all methods, implying the necessary boosting of emotions in helpfulness of reviews.\ BM25 can get more than 60\% helpfulness rate on both phone and laptop, whose outcome is slightly not as good as to GloVe\_$mSigmoid$ and GloVe\_$imSigmoid$ that combines descriptive sentences and consumer attitudes.\ The combination of GloVe and $iSigmoid$ is not stable in the two datasets, with an 0.65 helpfulness rate on reviews of phone but only 0.56 on that of laptop.\ Therefore GloVe\_$Sigmoid$ is the best ranking method to extract the most useful reviews given certain attributes for sellers and producers in e-commerce. 
\begin{table}[]
\caption{Helpfulness Rates of Different Methods for products in types of Phone and Laptop}
\centering
\begin{tabular}{lcccc}
\hline
                             &  & phone &  & laptop \\ \hline
BM25                         &  & 0.64  &  & 0.62   \\
GloVe                        &  & 0.48  &  & 0.40   \\
GloVe\_$Sigmoid$               &  & 0.72  &  & 0.70   \\
GloVe\_$iSigmoid$        &  & 0.63  &  & 0.56   \\
GloVe\_$mSigmoid$        &  & 0.65  &  & 0.69   \\
\textbf{GloVe\_$\bm{imSigmoid}$} &  & \textbf{0.69}  &  & \textbf{0.68}   \\ \hline
\end{tabular}
\label{helfulnessrate}
\end{table}

\begin{figure}[ht] 
    \begin{minipage}[t]{0.5\linewidth}
    \centering 
    \includegraphics[scale=1, width=1\textwidth]{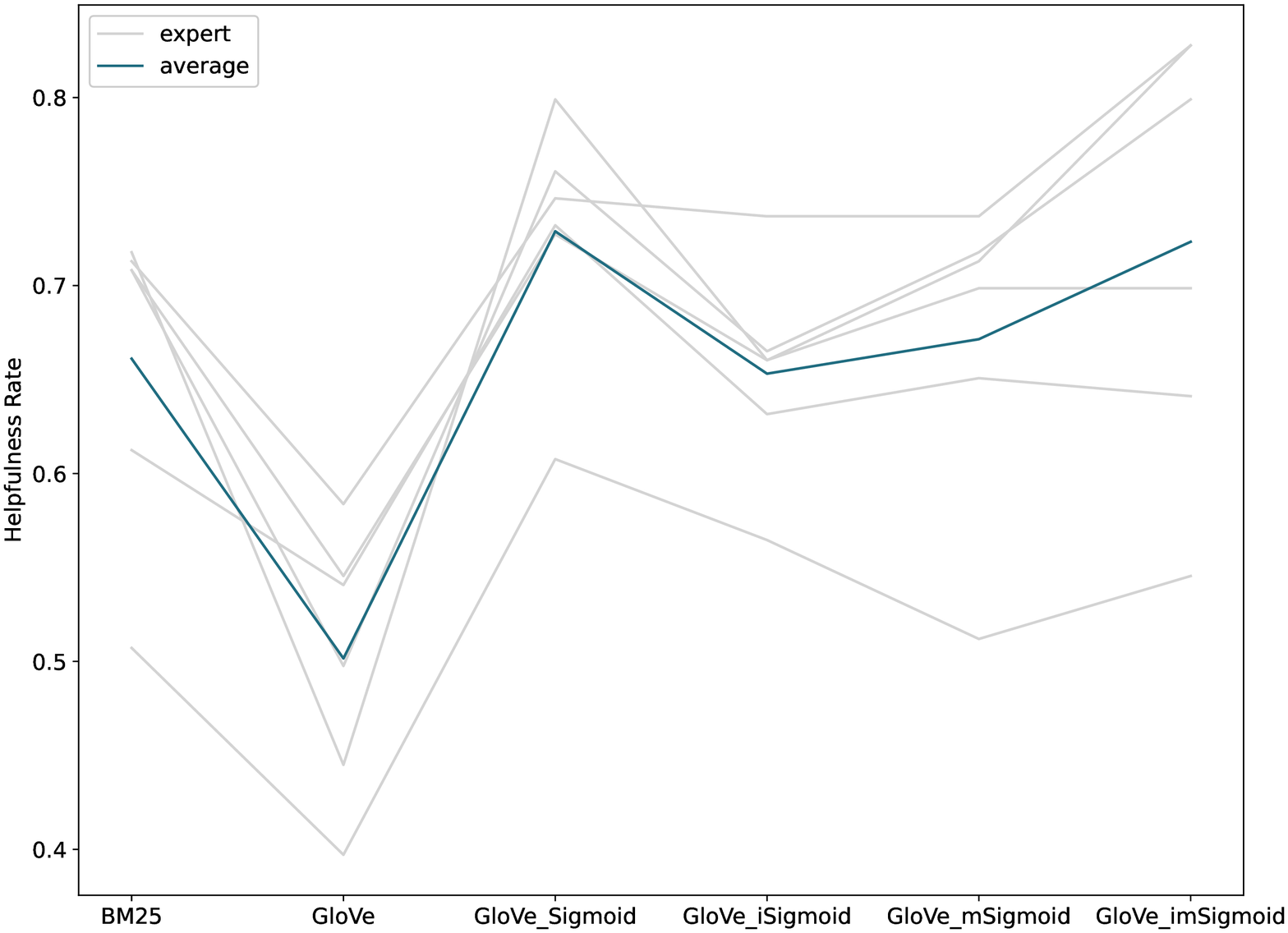} 
    \caption*{(a)phone}
    \end{minipage} 
    \begin{minipage}[t]{0.5\linewidth}
    \centering 
    \includegraphics[scale=1, width=1\textwidth]{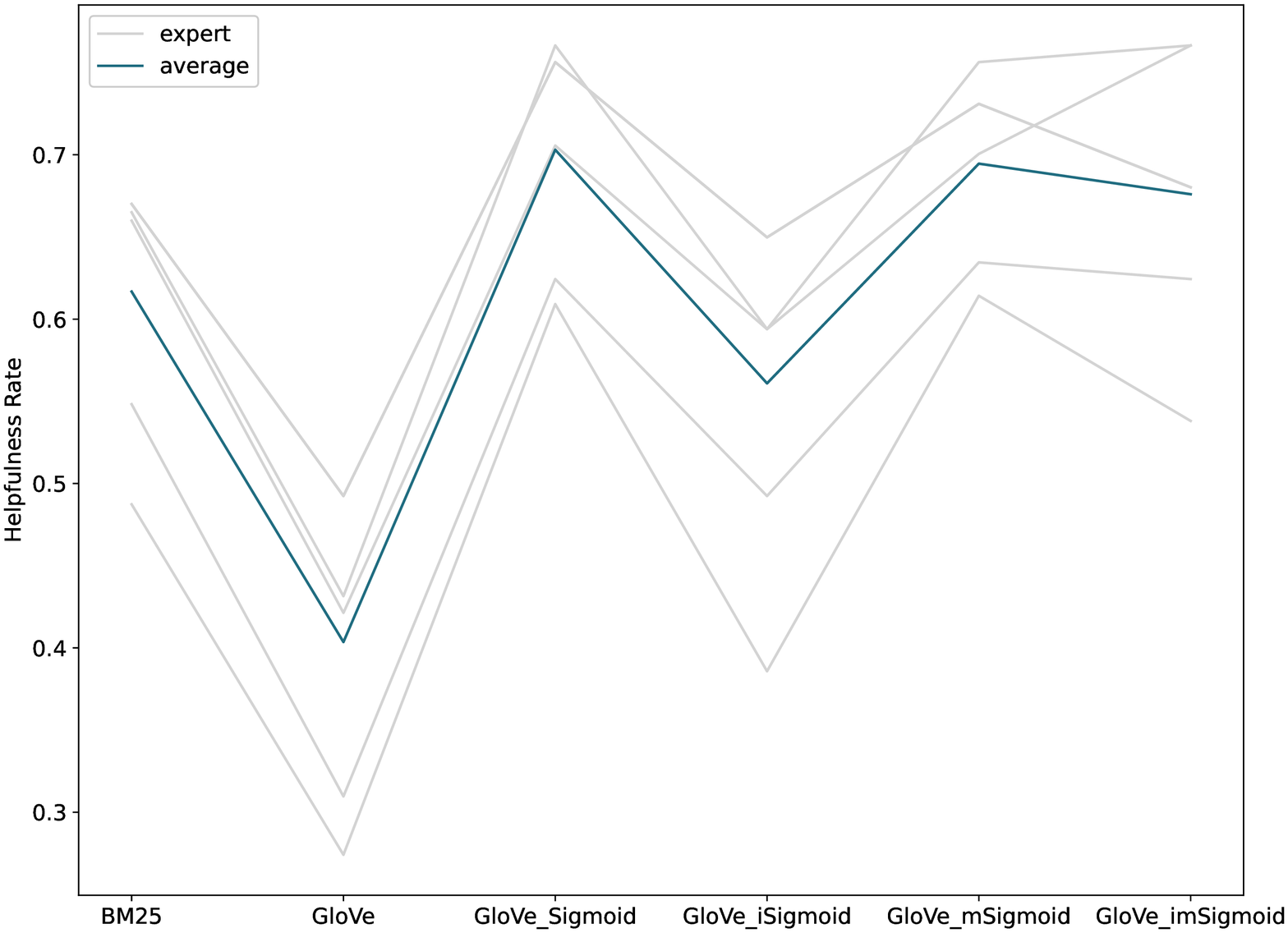} 
    \caption*{(b)laptop} 
    \end{minipage} 
    \caption{Average Helpfulness Rate of GloVe-based Methods and BM25}
    \label{fig-result_withexperts}
\end{figure}

\qquad As also can be seen in Figure \ref{fig-result_withexperts}, every experts' marked helpfulness rate and the average level are further demonstrated to testify the significance of performance comparison over ranking methods.\ It is again confirmed that GloVe\_$Sigmoid$ is consistently the best one among the baselines and variations.\ While GloVe\_$iSigmoid$ is the worst one in the variations.\ However, comparing with the baselines, BM25 and GloVe, all variations are better than them, at least not worse than them, meaning considering emotion polarity do help the ranking of helpfulness.\ With respect to negative reviews of laptop, GloVe\_$mSigmoid$ is ranked second but ranked third in those of phone, with a huge gap to the best method, GloVe\_$Sigmoid$, implying its unstable performance across product types. In the contrary, the stable outperformance of GloVe\_$Sigmoid$ suggests its competent capability of product independence in ranking negative reviews.

\subsection{Discussions}
\begin{figure}[ht] 
    \begin{minipage}[t]{0.5\linewidth}
    \centering 
    \includegraphics[scale=1, width=1\textwidth]{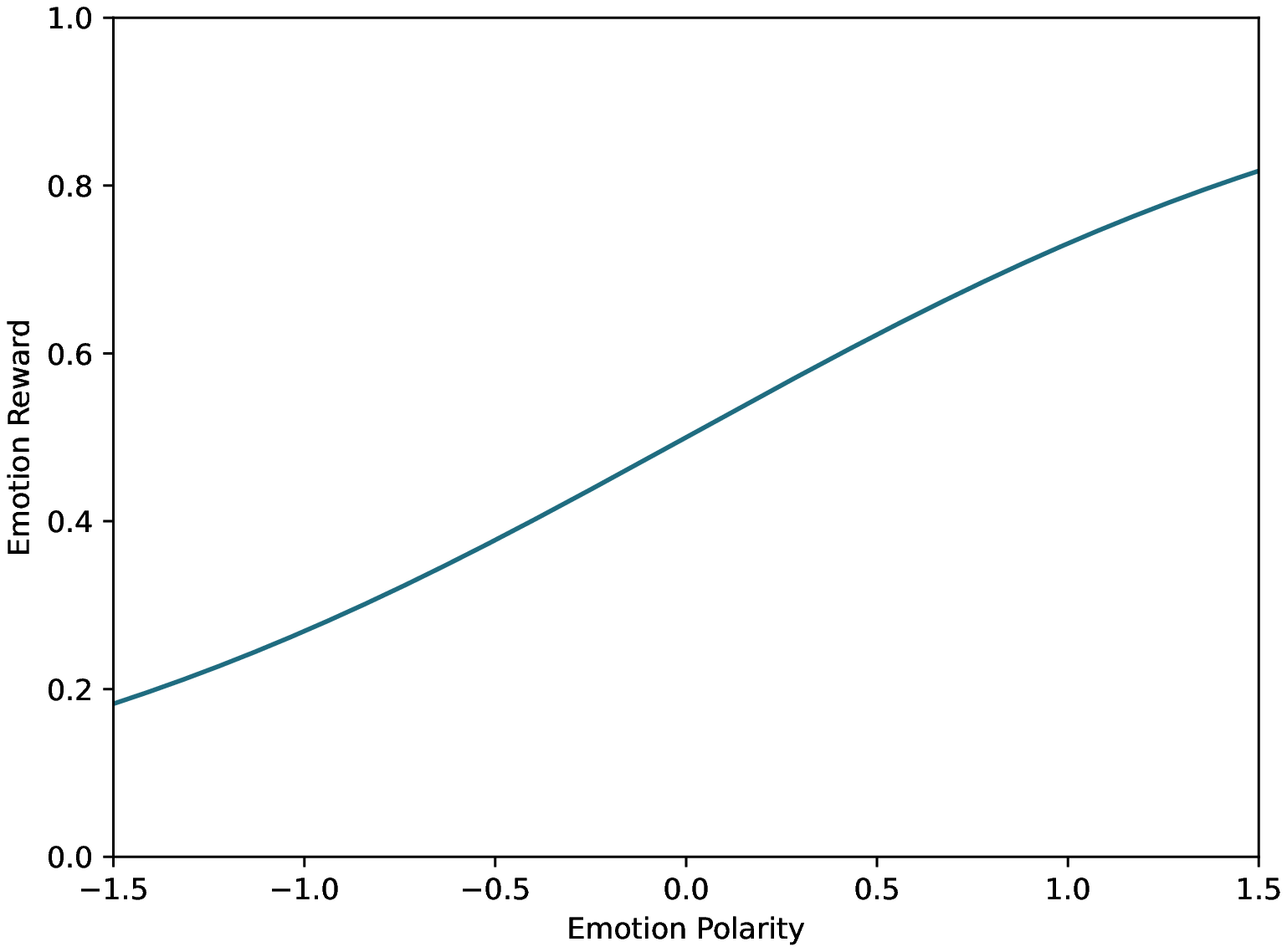} 
    \caption*{(a)$Sigmoid$}
    \end{minipage} 
    \begin{minipage}[t]{0.5\linewidth}
    \centering 
    \includegraphics[scale=1, width=1\textwidth]{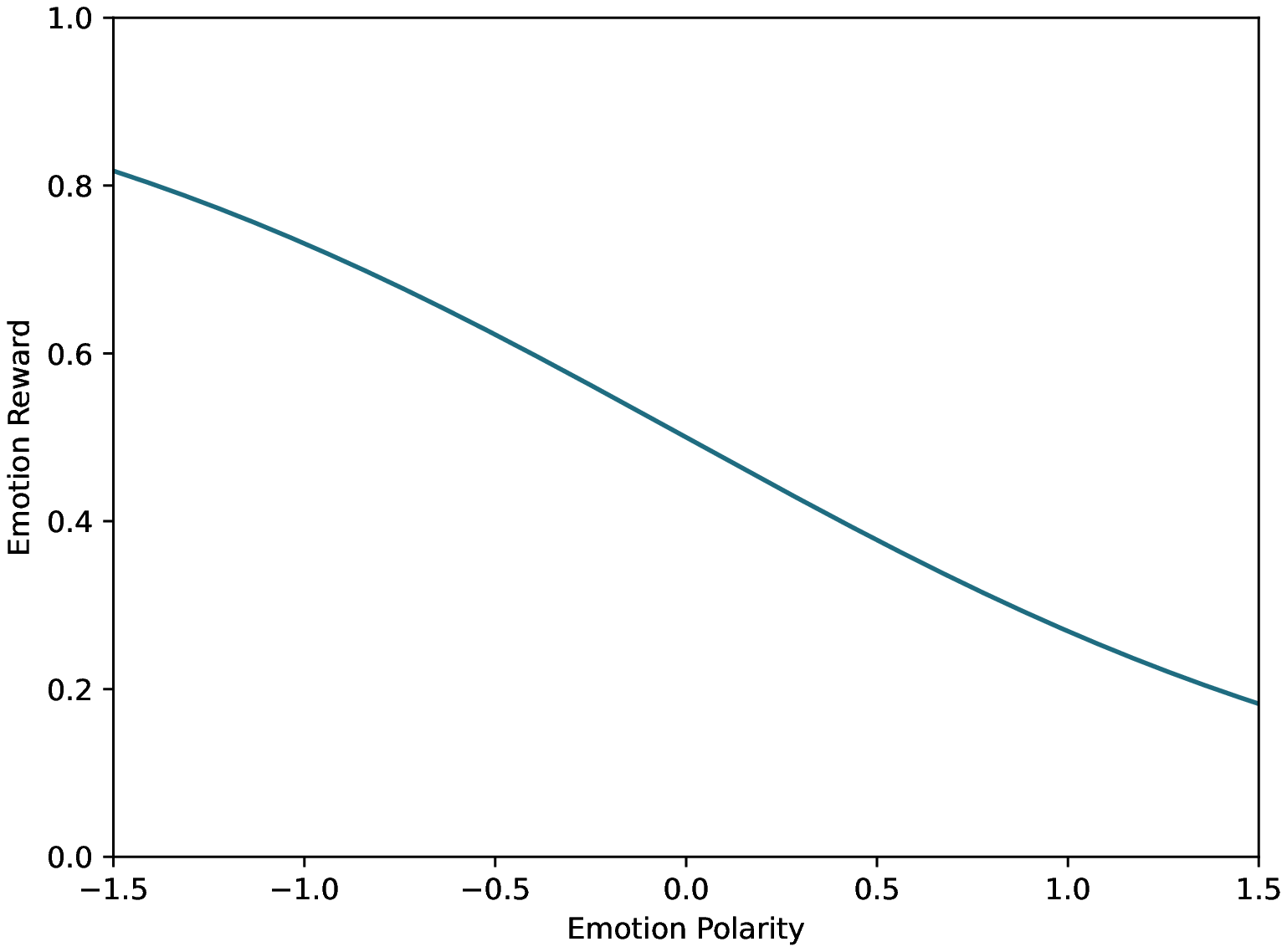} 
    \caption*{(b)$iSigmoid$} 
    \end{minipage}
    \begin{minipage}[t]{0.5\linewidth}
    \centering 
    \includegraphics[scale=1, width=1\textwidth]{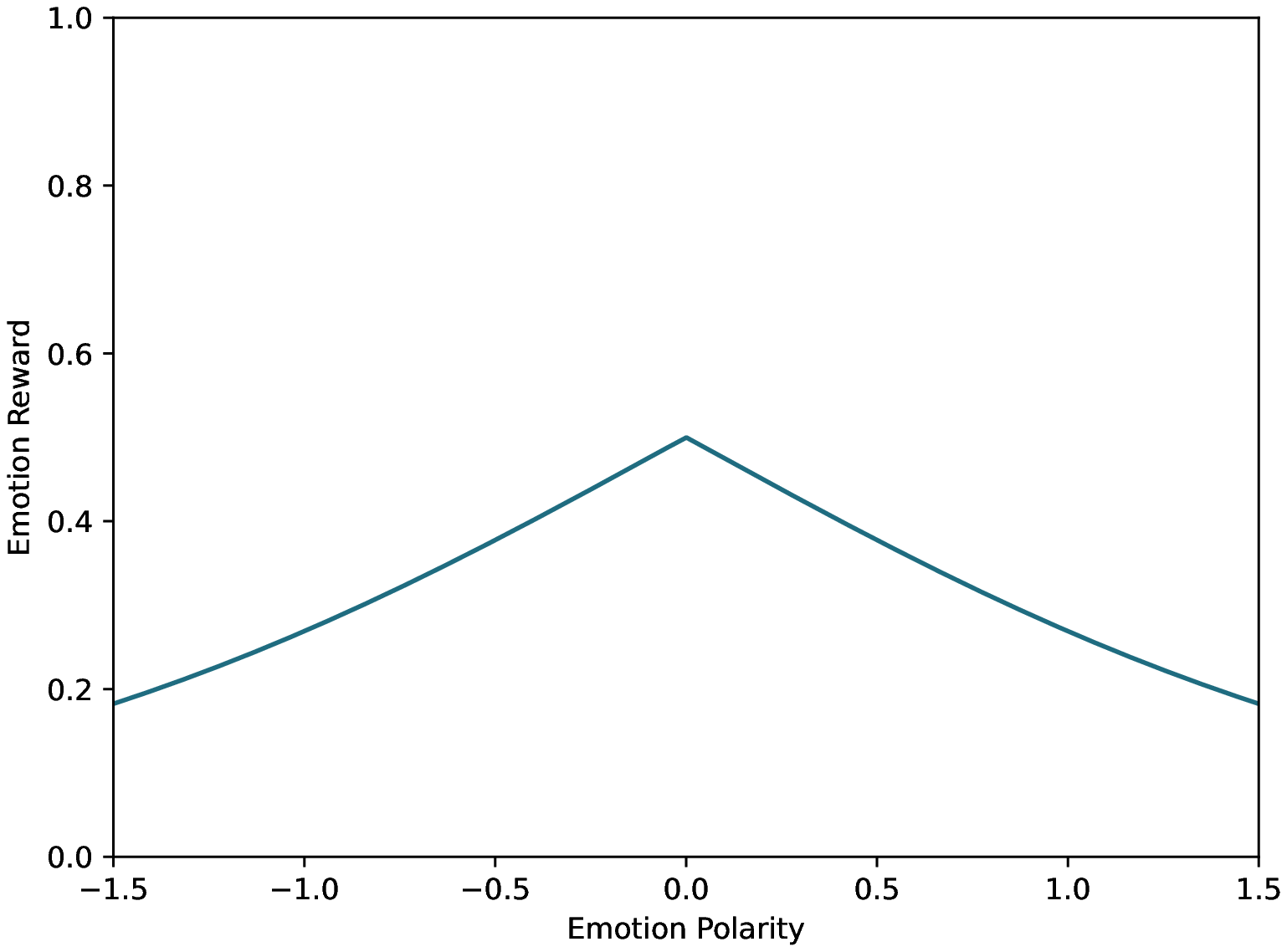} 
    \caption*{(c)$mSigmoid$}
    \end{minipage} 
    \begin{minipage}[t]{0.5\linewidth}
    \centering 
    \includegraphics[scale=1, width=1\textwidth]{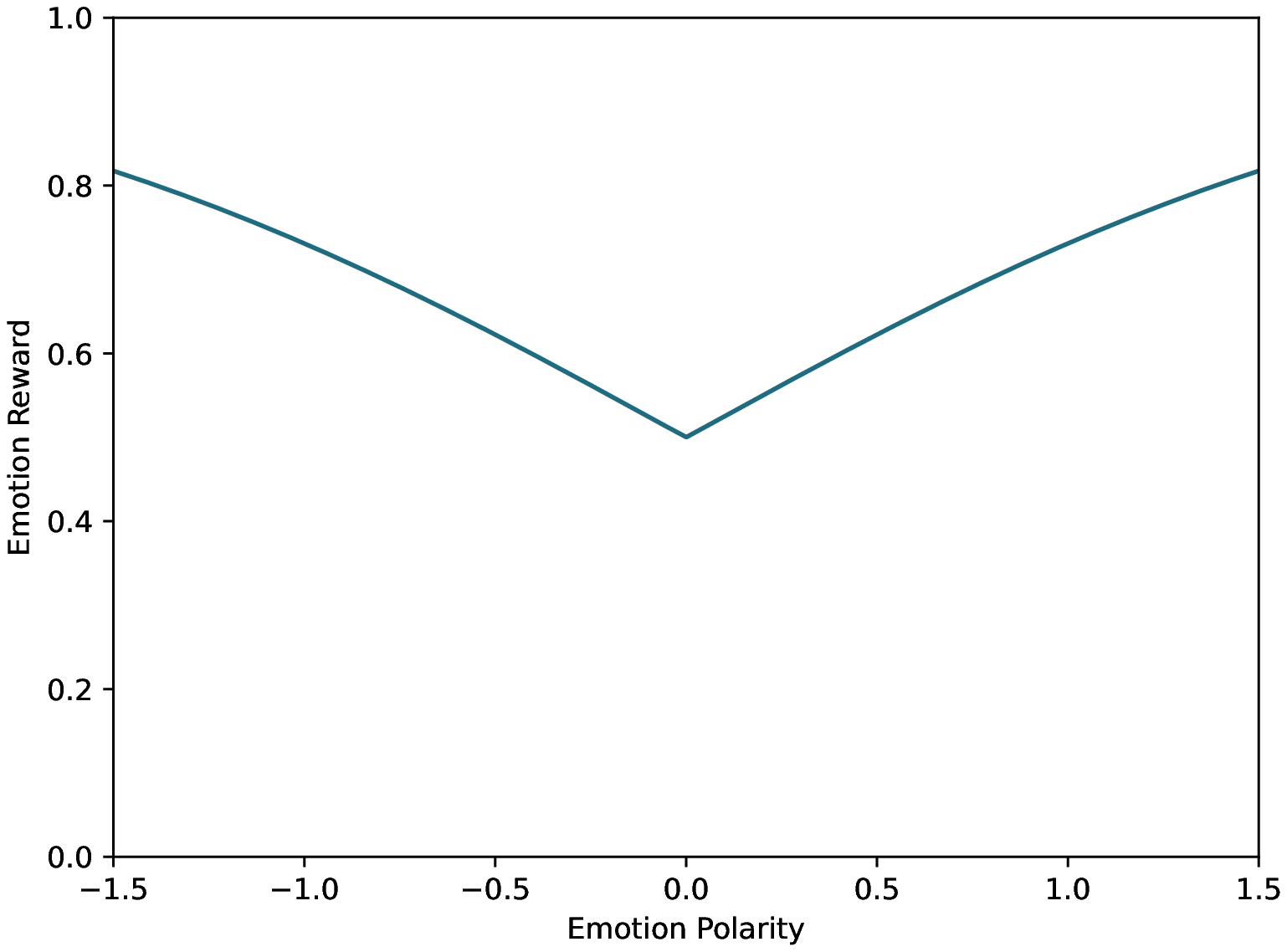} 
    \caption*{(d)$imSigmoid$}
    \end{minipage} 
    \caption{Emotion Reward of Each Method}
    \label{fig-sigmoids}
\end{figure}

\qquad To fully picture the emotion contributions in the ranking methods, in particular to profile the possible discrimination between positive and negative polarities, we further investigate how $e_{c}$ (see Eq.~\ref{eq:rankscore}) is tuned with the growth of emotion polarity in the ranking.\ We manually divided the GloVe\_$Sigmoid$ and its three variations into two groups according to the rewards, i.e., the adjusted value of positive or negative emotion polarity through Sigmoid functions.\ Figure \ref{fig-sigmoids} illustrates how the emotion rewards of each method fluctuate with growing emotion polarities.\ Here, we make two comparisons.\ In the first comparison, we group GloVe\_$Sigmoid$ and GloVe\_$imSigmoid$ as pos-high-reward, with GloVe\_$iSigmoid$ and GloVe\_$mSigmoid$ been grouped as pos-low-reward.\ The former group's helpfulness rates are greater than or not less than that of the latter group.\ Hence, the positive emotion existing in online reviews do help to increase the helpfulness rate of the ranking method.\ Besides, the fluctuations in pos-low-reward group are far more significant than the other group.\ In another word, the positive emotions are of high and stable value for e-commerce business in enhancing helpfulness of negative reviews. Within the group of pos-high-reward, the better performance of GloVe\_$Sigmoid$ than that of GloVe\_$imSigmoid$ further suggests that weighing heavily on negative emotions, in particular these extremely polarized ones (see Figure~\ref{fig-sigmoids}(d)), will undermine the helpfulness of negative reviews. In the contrary, as can be seen in Figure~\ref{fig-sigmoids}(a), suppressing the weights of negative polarities but promoting the rewards of positive counterparts will effectively enhance the helpfulness of negative reviews.

\qquad In a similar way, we also regard GloVe\_$iSigmoid$ and GloVe\_$imSigmoid$ as neg-high-reward group, and the rest of methods are the neg-low-reward group.\ Through the comparison of the neg-high-reward and the neg-low-reward, although the outcomes' stability of the neg-high-reward are not as good as that of the neg-low-reward, its performance sometimes greater than BM25.\ In this case, the value of negative emotions can not be simply ignored, especially as compared to solutions purely based on descriptive sentences.\ Given previous researches, we conclude that the negative emotions existing in negative reviews still have positive impacts in ranking reviews from the perspective of e-commerce keepers.\ Nevertheless, these impacts are not as consistent as the impacts brought by positive polarity for the significant fluctuations and should be well controlled to suppress the undermining from the extreme negative polarization.

It is widely thought that negative reviews are more helpful than positive reviews in e-commerce.\ By introducing emotion polarity to weigh the semantical similarity, we found that positive emotions do help rank negative reviews better than negative emotions do. It might credit to the rational narrative in poor ratings.\ Too much polarized negative reviews might be biased to of venting negative emotions instead of helpful feedback of product experience. It is also worthy noting that our conclusion contradicts with \cite{kim2020roles}. However, considering the unique setting in our evaluation that experts value the ranking from a perspective of sellers and producers instead of consumers, it is argued that this contradiction can be explained by the different context in the present study.

Our results supplement the existing understandings of review ranking with significant implications. From the unique view of e-commerce keepers such as retailers and producers, we first document the evidence of positive emotions in improving helpfulness of negative reviews. We accordingly argue that roles of emotions in review ranking could be heavily dependent on contexts. Existing debates on impact of emotions on helpfulness of reviews might be dissolved by this context-dependence. How to appropriately leverage emotions expressed by consumers should be carefully considered according to the specified scenarios. By weighing emotion polarity through a $Sigmoid$ scheme, the present study combines semantical similarity and emotion rewards to offer an efficient and effective ranking solution. Instead of helping future buyers make purchase decisions, our method is implemented to extract insights from negative reviews that could enhance service and production for sellers and producers. More importantly, this method could be expanded to other products and e-commerce websites with certain adjustments.  

\section{Conclusion}
\qquad We proposed a ranking method aiming at finding the most helpful negative reviews given specific product attributes.\ Based on data from JD.com, we found that positive emotions are more beneficial than negative emotions in ranking negative reviews' helpfulness from the unique view of online retailers and producers. By the combination of descriptive sentences and emotion polarity, our method at lease increases the helpfulness rate by 8\% comparing with the baseline approach.\ Our presented method can provide most relevant and useful online negative reviews for e-commerce practitioners of sellers and producers, with which they could make improvements of their business to increase the competitiveness on certain attributes.\ 

\qquad However, this study was conducted under the data from JD.com, especially the negative reviews.\ Further research is therefore needed to validate the generalization of the ranking method in other platforms.\ Apart from this, the ranking methods' implementation was limited to the workload of evaluation. Further work may focus on the validation of whether positive emotions can consistently help rank online negative reviews in other products for e-commerce practitioners.

\section*{References}
%\bibliography{mybibfile}

\begin{thebibliography}{45}
\expandafter\ifx\csname natexlab\endcsname\relax\def\natexlab#1{#1}\fi
\providecommand{\url}[1]{\texttt{#1}}
\providecommand{\href}[2]{#2}
\providecommand{\path}[1]{#1}
\providecommand{\DOIprefix}{doi:}
\providecommand{\ArXivprefix}{arXiv:}
\providecommand{\URLprefix}{URL: }
\providecommand{\Pubmedprefix}{pmid:}
\providecommand{\doi}[1]{\href{http://dx.doi.org/#1}{\path{#1}}}
\providecommand{\Pubmed}[1]{\href{pmid:#1}{\path{#1}}}
\providecommand{\bibinfo}[2]{#2}
\ifx\xfnm\undefined \def\xfnm[#1]{\unskip,\space#1}\fi
%Type = Inproceedings
\bibitem[{Baccianella et~al.(2010)Baccianella, Esuli and
  Sebastiani}]{baccianella2010sentiwordnet}
\bibinfo{author}{Baccianella\xfnm[ S.]}, \bibinfo{author}{Esuli\xfnm[ A.]},
  \bibinfo{author}{Sebastiani\xfnm[ F.]}.
\newblock \bibinfo{title}{{S}enti{W}ord{N}et 3.0: An enhanced lexical resource
  for sentiment analysis and opinion mining}.
\newblock In: \bibinfo{booktitle}{Proceedings of the Seventh International
  Conference on Language Resources and Evaluation ({LREC}'10)}.
  \bibinfo{address}{Valletta, Malta}: \bibinfo{publisher}{European Language
  Resources Association (ELRA)}; \bibinfo{year}{2010}. .
%Type = Article
\bibitem[{Bach and Kim(2012)}]{bach2012online}
\bibinfo{author}{Bach\xfnm[ S.B.]}, \bibinfo{author}{Kim\xfnm[ S.]}.
\newblock \bibinfo{title}{Online consumer complaint behaviors: The dynamics of
  service failures, consumers' word of mouth, and organization-consumer
  relationships}.
\newblock \bibinfo{journal}{International Journal of Strategic Communication}
  \bibinfo{year}{2012};\bibinfo{volume}{6}(\bibinfo{number}{1}):\bibinfo{pages}{59--76}.
%Type = Book
\bibitem[{Baeza-Yates et~al.(1999)Baeza-Yates, Ribeiro-Neto
  et~al.}]{baeza1999modern}
\bibinfo{author}{Baeza-Yates\xfnm[ R.]}, \bibinfo{author}{Ribeiro-Neto\xfnm[
  B.]}, et~al.
\newblock \bibinfo{title}{Modern information retrieval}.
\newblock volume \bibinfo{volume}{463}.
\newblock \bibinfo{publisher}{ACM press New York}, \bibinfo{year}{1999}.
%Type = Article
\bibitem[{Basuroy et~al.(2003)Basuroy, Chatterjee and
  Ravid}]{basuroy2003critical}
\bibinfo{author}{Basuroy\xfnm[ S.]}, \bibinfo{author}{Chatterjee\xfnm[ S.]},
  \bibinfo{author}{Ravid\xfnm[ S.A.]}.
\newblock \bibinfo{title}{How critical are critical reviews? {{The}} box office
  effects of film critics, star power, and budgets}.
\newblock \bibinfo{journal}{Journal of Marketing}
  \bibinfo{year}{2003};\bibinfo{volume}{67}(\bibinfo{number}{4}):\bibinfo{pages}{103--117}.
%Type = Article
\bibitem[{Berger et~al.(2010)Berger, Sorensen and
  Rasmussen}]{berger2010positive}
\bibinfo{author}{Berger\xfnm[ J.]}, \bibinfo{author}{Sorensen\xfnm[ A.T.]},
  \bibinfo{author}{Rasmussen\xfnm[ S.J.]}.
\newblock \bibinfo{title}{Positive effects of negative publicity: {{When}}
  negative reviews increase sales}.
\newblock \bibinfo{journal}{Marketing Science}
  \bibinfo{year}{2010};\bibinfo{volume}{29}(\bibinfo{number}{5}):\bibinfo{pages}{815--827}.
%Type = Article
\bibitem[{Bonnet and Nandan(2011)}]{bonnet2011transform}
\bibinfo{author}{Bonnet\xfnm[ D.]}, \bibinfo{author}{Nandan\xfnm[ P.]}.
\newblock \bibinfo{title}{Transform to the power of digital: Digital
  transformation as a driver of corporate performance}.
\newblock \bibinfo{journal}{Report, Capgemini Consulting}
  \bibinfo{year}{2011};.
%Type = Techreport
\bibitem[{Bradley and Lang(1999)}]{bradley1999affective}
\bibinfo{author}{Bradley\xfnm[ M.M.]}, \bibinfo{author}{Lang\xfnm[ P.J.]}.
\newblock \bibinfo{title}{Affective norms for English words (ANEW): Instruction
  manual and affective ratings}.
\newblock \bibinfo{type}{Technical Report}; Citeseer; \bibinfo{year}{1999}.
%Type = Article
\bibitem[{Cao et~al.(2011)Cao, Duan and Gan}]{cao2011exploring}
\bibinfo{author}{Cao\xfnm[ Q.]}, \bibinfo{author}{Duan\xfnm[ W.]},
  \bibinfo{author}{Gan\xfnm[ Q.]}.
\newblock \bibinfo{title}{Exploring determinants of voting for the
  ``helpfulness'' of online user reviews: {{A}} text mining approach}.
\newblock \bibinfo{journal}{Decision Support Systems}
  \bibinfo{year}{2011};\bibinfo{volume}{50}(\bibinfo{number}{2}):\bibinfo{pages}{511--521}.
%Type = Article
\bibitem[{Chen and Xie(2008)}]{chen2008online}
\bibinfo{author}{Chen\xfnm[ Y.]}, \bibinfo{author}{Xie\xfnm[ J.]}.
\newblock \bibinfo{title}{Online consumer review: {{Word}}-of-mouth as a new
  element of marketing communication mix}.
\newblock \bibinfo{journal}{Management Science}
  \bibinfo{year}{2008};\bibinfo{volume}{54}(\bibinfo{number}{3}):\bibinfo{pages}{477--491}.
%Type = Article
\bibitem[{Chen and Lurie(2013)}]{chen2013temporal}
\bibinfo{author}{Chen\xfnm[ Z.]}, \bibinfo{author}{Lurie\xfnm[ N.H.]}.
\newblock \bibinfo{title}{Temporal contiguity and negativity bias in the impact
  of online word of mouth}.
\newblock \bibinfo{journal}{Journal of Marketing Research}
  \bibinfo{year}{2013};\bibinfo{volume}{50}(\bibinfo{number}{4}):\bibinfo{pages}{463--476}.
\newblock \bibinfo{note}{00215}.
%Type = Article
\bibitem[{Chevalier and Mayzlin(2006)}]{chevalier2006effect}
\bibinfo{author}{Chevalier\xfnm[ J.A.]}, \bibinfo{author}{Mayzlin\xfnm[ D.]}.
\newblock \bibinfo{title}{The effect of word of mouth on sales: Online book
  reviews}.
\newblock \bibinfo{journal}{Journal of Marketing Research}
  \bibinfo{year}{2006};\bibinfo{volume}{43}(\bibinfo{number}{3}):\bibinfo{pages}{345--354}.
%Type = Article
\bibitem[{De~Matos and Rossi(2008)}]{de2008word}
\bibinfo{author}{De~Matos\xfnm[ C.A.]}, \bibinfo{author}{Rossi\xfnm[ C.A.V.]}.
\newblock \bibinfo{title}{Word-of-mouth communications in marketing: a
  meta-analytic review of the antecedents and moderators}.
\newblock \bibinfo{journal}{Journal of the Academy of Marketing Science}
  \bibinfo{year}{2008};\bibinfo{volume}{36}(\bibinfo{number}{4}):\bibinfo{pages}{578--596}.
%Type = Article
\bibitem[{Dellarocas(2003)}]{dellarocas2003digitization}
\bibinfo{author}{Dellarocas\xfnm[ C.]}.
\newblock \bibinfo{title}{The digitization of word of mouth: {{Promise}} and
  challenges of online feedback mechanisms}.
\newblock \bibinfo{journal}{Management science}
  \bibinfo{year}{2003};\bibinfo{volume}{49}(\bibinfo{number}{10}):\bibinfo{pages}{1407--1424}.
\newblock \bibinfo{note}{03586}.
%Type = Article
\bibitem[{Devlin et~al.(2018)Devlin, Chang, Lee and Toutanova}]{devlin2018bert}
\bibinfo{author}{Devlin\xfnm[ J.]}, \bibinfo{author}{Chang\xfnm[ M.W.]},
  \bibinfo{author}{Lee\xfnm[ K.]}, \bibinfo{author}{Toutanova\xfnm[ K.]}.
\newblock \bibinfo{title}{Bert: Pre-training of deep bidirectional transformers
  for language understanding}.
\newblock \bibinfo{journal}{arXiv:181004805} \bibinfo{year}{2018};.
%Type = Article
\bibitem[{Duan et~al.(2008)Duan, Gu and Whinston}]{duan2008online}
\bibinfo{author}{Duan\xfnm[ W.]}, \bibinfo{author}{Gu\xfnm[ B.]},
  \bibinfo{author}{Whinston\xfnm[ A.B.]}.
\newblock \bibinfo{title}{Do online reviews matter? \textemdash{} {{An}}
  empirical investigation of panel data}.
\newblock \bibinfo{journal}{Decision Support Systems}
  \bibinfo{year}{2008};\bibinfo{volume}{45}(\bibinfo{number}{4}):\bibinfo{pages}{1007--1016}.
\newblock \bibinfo{note}{01614}.
%Type = Article
\bibitem[{Forman et~al.(2008)Forman, Ghose and
  Wiesenfeld}]{forman2008examining}
\bibinfo{author}{Forman\xfnm[ C.]}, \bibinfo{author}{Ghose\xfnm[ A.]},
  \bibinfo{author}{Wiesenfeld\xfnm[ B.]}.
\newblock \bibinfo{title}{Examining the relationship between reviews and sales:
  {{The}} role of reviewer identity disclosure in electronic markets}.
\newblock \bibinfo{journal}{Information Systems Research}
  \bibinfo{year}{2008};\bibinfo{volume}{19}(\bibinfo{number}{3}):\bibinfo{pages}{291--313}.
\newblock \bibinfo{note}{01252}.
%Type = Article
\bibitem[{Ghose and Ipeirotis(2011)}]{ghose2011estimating}
\bibinfo{author}{Ghose\xfnm[ A.]}, \bibinfo{author}{Ipeirotis\xfnm[ P.G.]}.
\newblock \bibinfo{title}{Estimating the helpfulness and economic impact of
  product reviews: {{Mining}} text and reviewer characteristics}.
\newblock \bibinfo{journal}{IEEE Transactions on Knowledge and Data
  Engineering}
  \bibinfo{year}{2011};\bibinfo{volume}{23}(\bibinfo{number}{10}):\bibinfo{pages}{1498--1512}.
%Type = Inproceedings
\bibitem[{Hsu et~al.(2009)Hsu, Khabiri and Caverlee}]{hsu2009ranking}
\bibinfo{author}{Hsu\xfnm[ C.F.]}, \bibinfo{author}{Khabiri\xfnm[ E.]},
  \bibinfo{author}{Caverlee\xfnm[ J.]}.
\newblock \bibinfo{title}{Ranking {{Comments}} on the {{Social Web}}}.
\newblock In: \bibinfo{booktitle}{2009 {{International Conference}} on
  {{Computational Science}} and {{Engineering}}}. \bibinfo{address}{Vancouver,
  BC, Canada}: \bibinfo{publisher}{{IEEE}}; \bibinfo{year}{2009}. p.
  \bibinfo{pages}{90--97}.
\newblock \bibinfo{note}{00124}.
%Type = Article
\bibitem[{Huang et~al.(2017)Huang, Hong and Burtch}]{huang2017social}
\bibinfo{author}{Huang\xfnm[ N.]}, \bibinfo{author}{Hong\xfnm[ Y.]},
  \bibinfo{author}{Burtch\xfnm[ G.]}.
\newblock \bibinfo{title}{Social network integration and user content
  generation: {{Evidence}} from natural experiments}.
\newblock \bibinfo{journal}{MIS Quarterly}
  \bibinfo{year}{2017};\bibinfo{volume}{41}(\bibinfo{number}{4}):\bibinfo{pages}{1035--1058}.
%Type = Article
\bibitem[{Kim and Hwang(2020)}]{kim2020roles}
\bibinfo{author}{Kim\xfnm[ J.M.]}, \bibinfo{author}{Hwang\xfnm[ K.]}.
\newblock \bibinfo{title}{Roles of emotional expressions in review consumption
  and generation processes}.
\newblock \bibinfo{journal}{International Journal of Hospitality Management}
  \bibinfo{year}{2020};\bibinfo{volume}{86}:\bibinfo{pages}{102454}.
%Type = Inproceedings
\bibitem[{Kusner et~al.(2015)Kusner, Sun, Kolkin and
  Weinberger}]{kusnerWordEmbeddingsDocument2015}
\bibinfo{author}{Kusner\xfnm[ M.J.]}, \bibinfo{author}{Sun\xfnm[ Y.]},
  \bibinfo{author}{Kolkin\xfnm[ N.I.]}, \bibinfo{author}{Weinberger\xfnm[
  K.Q.]}.
\newblock \bibinfo{title}{From word embeddings to document distances}.
\newblock In: \bibinfo{booktitle}{Proceedings of the 32nd International
  Conference on International Conference on Machine Learning - Volume 37}.
  \bibinfo{publisher}{JMLR.org}; ICML’15; \bibinfo{year}{2015}. p.
  \bibinfo{pages}{957–966}.
%Type = Inproceedings
\bibitem[{Le and Mikolov(2014)}]{le2014distributed}
\bibinfo{author}{Le\xfnm[ Q.]}, \bibinfo{author}{Mikolov\xfnm[ T.]}.
\newblock \bibinfo{title}{Distributed representations of sentences and
  documents}.
\newblock In: \bibinfo{booktitle}{Proceedings of the 31st International
  Conference on International Conference on Machine Learning - Volume 32}.
  \bibinfo{publisher}{JMLR.org}; ICML’14; \bibinfo{year}{2014}. p.
  \bibinfo{pages}{II–1188–II–1196}.
%Type = Article
\bibitem[{Lee et~al.(2017)Lee, Jeong and Lee}]{lee2017roles}
\bibinfo{author}{Lee\xfnm[ M.]}, \bibinfo{author}{Jeong\xfnm[ M.]},
  \bibinfo{author}{Lee\xfnm[ J.]}.
\newblock \bibinfo{title}{Roles of negative emotions in customers’ perceived
  helpfulness of hotel reviews on a user-generated review website}.
\newblock \bibinfo{journal}{International Journal of Contemporary Hospitality
  Management}
  \bibinfo{year}{2017};\bibinfo{volume}{29}(\bibinfo{number}{2}):\bibinfo{pages}{762--783}.
%Type = Inproceedings
\bibitem[{Liu et~al.(2008)Liu, Huang, An and Yu}]{liu2008modeling}
\bibinfo{author}{Liu\xfnm[ Y.]}, \bibinfo{author}{Huang\xfnm[ X.]},
  \bibinfo{author}{An\xfnm[ A.]}, \bibinfo{author}{Yu\xfnm[ X.]}.
\newblock \bibinfo{title}{Modeling and {{Predicting}} the {{Helpfulness}} of
  {{Online Reviews}}}.
\newblock In: \bibinfo{booktitle}{2008 {{Eighth IEEE International Conference}}
  on {{Data Mining}}}. \bibinfo{address}{Pisa, Italy}:
  \bibinfo{publisher}{{IEEE}}; \bibinfo{year}{2008}. p.
  \bibinfo{pages}{443--452}.
\newblock \bibinfo{note}{00230}.
%Type = Article
\bibitem[{Ludwig et~al.(2013)Ludwig, {\noopsort{ruyter}}{de Ruyter}, Friedman,
  Br{\"u}ggen, Wetzels and Pfann}]{ludwig2013more}
\bibinfo{author}{Ludwig\xfnm[ S.]}, \bibinfo{author}{{\noopsort{ruyter}}{de
  Ruyter}\xfnm[ K.]}, \bibinfo{author}{Friedman\xfnm[ M.]},
  \bibinfo{author}{Br{\"u}ggen\xfnm[ E.C.]}, \bibinfo{author}{Wetzels\xfnm[
  M.]}, \bibinfo{author}{Pfann\xfnm[ G.]}.
\newblock \bibinfo{title}{More than {{Words}}: {{The Influence}} of {{Affective
  Content}} and {{Linguistic Style Matches}} in {{Online Reviews}} on
  {{Conversion Rates}}}.
\newblock \bibinfo{journal}{Journal of Marketing}
  \bibinfo{year}{2013};\bibinfo{volume}{77}(\bibinfo{number}{1}):\bibinfo{pages}{87--103}.
\newblock \bibinfo{note}{00336}.
%Type = Article
\bibitem[{Luo(2009)}]{luo2009quantifying}
\bibinfo{author}{Luo\xfnm[ X.]}.
\newblock \bibinfo{title}{Quantifying the long-term impact of negative word of
  mouth on cash flows and stock prices}.
\newblock \bibinfo{journal}{Marketing Science}
  \bibinfo{year}{2009};\bibinfo{volume}{28}(\bibinfo{number}{1}):\bibinfo{pages}{148--165}.
%Type = Inproceedings
\bibitem[{Martin and Pu(2014)}]{martin2014prediction}
\bibinfo{author}{Martin\xfnm[ L.]}, \bibinfo{author}{Pu\xfnm[ P.]}.
\newblock \bibinfo{title}{Prediction of helpful reviews using emotions
  extraction}.
\newblock In: \bibinfo{booktitle}{Proceedings of the Twenty-Eighth AAAI
  Conference on Artificial Intelligence}. \bibinfo{publisher}{AAAI Press};
  AAAI’14; \bibinfo{year}{2014}. p. \bibinfo{pages}{1551–--1557}.
%Type = Article
\bibitem[{Mikolov et~al.(2013{\natexlab{a}})Mikolov, Chen, Corrado and
  Dean}]{mikolov2013efficient}
\bibinfo{author}{Mikolov\xfnm[ T.]}, \bibinfo{author}{Chen\xfnm[ K.]},
  \bibinfo{author}{Corrado\xfnm[ G.]}, \bibinfo{author}{Dean\xfnm[ J.]}.
\newblock \bibinfo{title}{Efficient estimation of word representations in
  vector space}.
\newblock \bibinfo{journal}{arXiv:13013781}
  \bibinfo{year}{2013}{\natexlab{a}};.
%Type = Inproceedings
\bibitem[{Mikolov et~al.(2013{\natexlab{b}})Mikolov, Sutskever, Chen, Corrado
  and Dean}]{mikolov2013distributed}
\bibinfo{author}{Mikolov\xfnm[ T.]}, \bibinfo{author}{Sutskever\xfnm[ I.]},
  \bibinfo{author}{Chen\xfnm[ K.]}, \bibinfo{author}{Corrado\xfnm[ G.S.]},
  \bibinfo{author}{Dean\xfnm[ J.]}.
\newblock \bibinfo{title}{Distributed representations of words and phrases and
  their compositionality}.
\newblock In: \bibinfo{booktitle}{Advances in neural information processing
  systems}. \bibinfo{year}{2013}{\natexlab{b}}. p. \bibinfo{pages}{3111--3119}.
%Type = Article
\bibitem[{Moe and Trusov(2011)}]{moe2011value}
\bibinfo{author}{Moe\xfnm[ W.W.]}, \bibinfo{author}{Trusov\xfnm[ M.]}.
\newblock \bibinfo{title}{The value of social dynamics in online product
  ratings forums}.
\newblock \bibinfo{journal}{Journal of Marketing Research}
  \bibinfo{year}{2011};\bibinfo{volume}{48}(\bibinfo{number}{3}):\bibinfo{pages}{444--456}.
%Type = Article
\bibitem[{Mudambi and Schuff(2010)}]{mudambi2010makes}
\bibinfo{author}{Mudambi\xfnm[ S.M.]}, \bibinfo{author}{Schuff\xfnm[ D.]}.
\newblock \bibinfo{title}{What makes a helpful review? {{A}} study of customer
  reviews on {{Amazon}}. com}.
\newblock \bibinfo{journal}{MIS Quarterly}
  \bibinfo{year}{2010};\bibinfo{volume}{34}(\bibinfo{number}{1}):\bibinfo{pages}{185--200}.
%Type = Article
\bibitem[{Pennebaker et~al.(2003)Pennebaker, Mehl and
  Niederhoffer}]{pennebaker2003psychological}
\bibinfo{author}{Pennebaker\xfnm[ J.W.]}, \bibinfo{author}{Mehl\xfnm[ M.R.]},
  \bibinfo{author}{Niederhoffer\xfnm[ K.G.]}.
\newblock \bibinfo{title}{Psychological aspects of natural language use: Our
  words, our selves}.
\newblock \bibinfo{journal}{Annual Review of Psychology}
  \bibinfo{year}{2003};\bibinfo{volume}{54}(\bibinfo{number}{1}):\bibinfo{pages}{547--577}.
%Type = Inproceedings
\bibitem[{Pennington et~al.(2014)Pennington, Socher and
  Manning}]{pennington2014glove}
\bibinfo{author}{Pennington\xfnm[ J.]}, \bibinfo{author}{Socher\xfnm[ R.]},
  \bibinfo{author}{Manning\xfnm[ C.]}.
\newblock \bibinfo{title}{Glove: Global vectors for word representation}.
\newblock In: \bibinfo{booktitle}{Proceedings of the 2014 conference on
  empirical methods in natural language processing (EMNLP)}.
  \bibinfo{year}{2014}. p. \bibinfo{pages}{1532--1543}.
%Type = Article
\bibitem[{Perkins(2009)}]{perkins2009power}
\bibinfo{author}{Perkins\xfnm[ B.]}.
\newblock \bibinfo{title}{The power of viral revenge}.
\newblock \bibinfo{journal}{Computerworld}
  \bibinfo{year}{2009};\bibinfo{volume}{40}:\bibinfo{pages}{2--3}.
%Type = Inproceedings
\bibitem[{Robertson et~al.(1995)Robertson, Walker, Jones, Hancock-Beaulieu and
  Gatford}]{robertson1995okapi}
\bibinfo{author}{Robertson\xfnm[ S.]}, \bibinfo{author}{Walker\xfnm[ S.]},
  \bibinfo{author}{Jones\xfnm[ S.]}, \bibinfo{author}{Hancock-Beaulieu\xfnm[
  M.M.]}, \bibinfo{author}{Gatford\xfnm[ M.]}.
\newblock \bibinfo{title}{Okapi at trec-3}.
\newblock In: \bibinfo{booktitle}{Overview of the Third Text REtrieval
  Conference (TREC-3)}. \bibinfo{publisher}{Gaithersburg, MD: NIST};
  \bibinfo{year}{1995}. p. \bibinfo{pages}{109--126}.
%Type = Inproceedings
\bibitem[{Robertson and Walker(1994)}]{robertson1994some}
\bibinfo{author}{Robertson\xfnm[ S.E.]}, \bibinfo{author}{Walker\xfnm[ S.]}.
\newblock \bibinfo{title}{Some simple effective approximations to the 2-poisson
  model for probabilistic weighted retrieval}.
\newblock In: \bibinfo{booktitle}{{{SIGIR}}'94}.
  \bibinfo{organization}{{Springer}}; \bibinfo{year}{1994}. p.
  \bibinfo{pages}{232--241}.
%Type = Article
\bibitem[{Saumya et~al.(2018)Saumya, Singh, Baabdullah, Rana and
  Dwivedi}]{saumya2018ranking}
\bibinfo{author}{Saumya\xfnm[ S.]}, \bibinfo{author}{Singh\xfnm[ J.P.]},
  \bibinfo{author}{Baabdullah\xfnm[ A.M.]}, \bibinfo{author}{Rana\xfnm[ N.P.]},
  \bibinfo{author}{Dwivedi\xfnm[ Y.K.]}.
\newblock \bibinfo{title}{Ranking online consumer reviews}.
\newblock \bibinfo{journal}{Electronic Commerce Research and Applications}
  \bibinfo{year}{2018};\bibinfo{volume}{29}:\bibinfo{pages}{78--89}.
%Type = Article
\bibitem[{Schindler and Bickart(2012)}]{schindler2012perceived}
\bibinfo{author}{Schindler\xfnm[ R.M.]}, \bibinfo{author}{Bickart\xfnm[ B.]}.
\newblock \bibinfo{title}{Perceived helpfulness of online consumer reviews:
  {{The}} role of message content and style}.
\newblock \bibinfo{journal}{Journal of Consumer Behaviour}
  \bibinfo{year}{2012};\bibinfo{volume}{11}(\bibinfo{number}{3}):\bibinfo{pages}{234--243}.
\newblock \bibinfo{note}{00196}.
%Type = Article
\bibitem[{Sweeney et~al.(2014)Sweeney, Soutar and
  Mazzarol}]{sweeney2014factors}
\bibinfo{author}{Sweeney\xfnm[ J.]}, \bibinfo{author}{Soutar\xfnm[ G.]},
  \bibinfo{author}{Mazzarol\xfnm[ T.]}.
\newblock \bibinfo{title}{Factors enhancing word-of-mouth influence: positive
  and negative service-related messages}.
\newblock \bibinfo{journal}{European Journal of Marketing}
  \bibinfo{year}{2014};\bibinfo{volume}{48}(\bibinfo{number}{1/2}):\bibinfo{pages}{336--359}.
%Type = Article
\bibitem[{Sweeney et~al.(2012)Sweeney, Soutar and Mazzarol}]{sweeney2012word}
\bibinfo{author}{Sweeney\xfnm[ J.C.]}, \bibinfo{author}{Soutar\xfnm[ G.N.]},
  \bibinfo{author}{Mazzarol\xfnm[ T.]}.
\newblock \bibinfo{title}{Word of mouth: measuring the power of individual
  messages}.
\newblock \bibinfo{journal}{European Journal of Marketing}
  \bibinfo{year}{2012};\bibinfo{volume}{46}(\bibinfo{number}{1/2}):\bibinfo{pages}{237--257}.
%Type = Article
\bibitem[{Verhagen et~al.(2013)Verhagen, Nauta and
  Feldberg}]{verhagen2013negative}
\bibinfo{author}{Verhagen\xfnm[ T.]}, \bibinfo{author}{Nauta\xfnm[ A.]},
  \bibinfo{author}{Feldberg\xfnm[ F.]}.
\newblock \bibinfo{title}{Negative online word-of-mouth: Behavioral indicator
  or emotional release?}
\newblock \bibinfo{journal}{Computers in Human Behavior}
  \bibinfo{year}{2013};\bibinfo{volume}{29}(\bibinfo{number}{4}):\bibinfo{pages}{1430--1440}.
%Type = Article
\bibitem[{Willemsen et~al.(2011)Willemsen, Neijens, Bronner and
  De~Ridder}]{willemsen2011highly}
\bibinfo{author}{Willemsen\xfnm[ L.M.]}, \bibinfo{author}{Neijens\xfnm[ P.C.]},
  \bibinfo{author}{Bronner\xfnm[ F.]}, \bibinfo{author}{De~Ridder\xfnm[ J.A.]}.
\newblock \bibinfo{title}{``{{Highly}} recommended!'' {{The}} content
  characteristics and perceived usefulness of online consumer reviews}.
\newblock \bibinfo{journal}{Journal of Computer-Mediated Communication}
  \bibinfo{year}{2011};\bibinfo{volume}{17}(\bibinfo{number}{1}):\bibinfo{pages}{19--38}.
%Type = Article
\bibitem[{Ye et~al.(2011)Ye, Law, Gu and Chen}]{ye2011influence}
\bibinfo{author}{Ye\xfnm[ Q.]}, \bibinfo{author}{Law\xfnm[ R.]},
  \bibinfo{author}{Gu\xfnm[ B.]}, \bibinfo{author}{Chen\xfnm[ W.]}.
\newblock \bibinfo{title}{The influence of user-generated content on traveler
  behavior: {{An}} empirical investigation on the effects of e-word-of-mouth to
  hotel online bookings}.
\newblock \bibinfo{journal}{Computers in Human Behavior}
  \bibinfo{year}{2011};\bibinfo{volume}{27}(\bibinfo{number}{2}):\bibinfo{pages}{634--639}.
%Type = Article
\bibitem[{Yin et~al.(2014)Yin, Bond and Zhang}]{yin2014anxious}
\bibinfo{author}{Yin\xfnm[ D.]}, \bibinfo{author}{Bond\xfnm[ S.D.]},
  \bibinfo{author}{Zhang\xfnm[ H.]}.
\newblock \bibinfo{title}{Anxious or {{Angry}}? {{Effects}} of {{Discrete
  Emotions}} on the {{Perceived Helpfulness}} of {{Online Reviews}}}.
\newblock \bibinfo{journal}{MIS Quarterly}
  \bibinfo{year}{2014};\bibinfo{volume}{38}(\bibinfo{number}{2}):\bibinfo{pages}{539--560}.
%Type = Article
\bibitem[{Yu et~al.(2019)Yu, Su and Luo}]{yu2019improving}
\bibinfo{author}{Yu\xfnm[ S.]}, \bibinfo{author}{Su\xfnm[ J.]},
  \bibinfo{author}{Luo\xfnm[ D.]}.
\newblock \bibinfo{title}{Improving {BERT}-based text classification with
  auxiliary sentence and domain knowledge}.
\newblock \bibinfo{journal}{IEEE Access}
  \bibinfo{year}{2019};\bibinfo{volume}{7}:\bibinfo{pages}{176600--176612}.

\end{thebibliography}
\newcommand{\noopsort}[1]{}

\end{document}